# Matlab Implementation of Machine Vision Algorithm on Ballast Degradation Evaluation

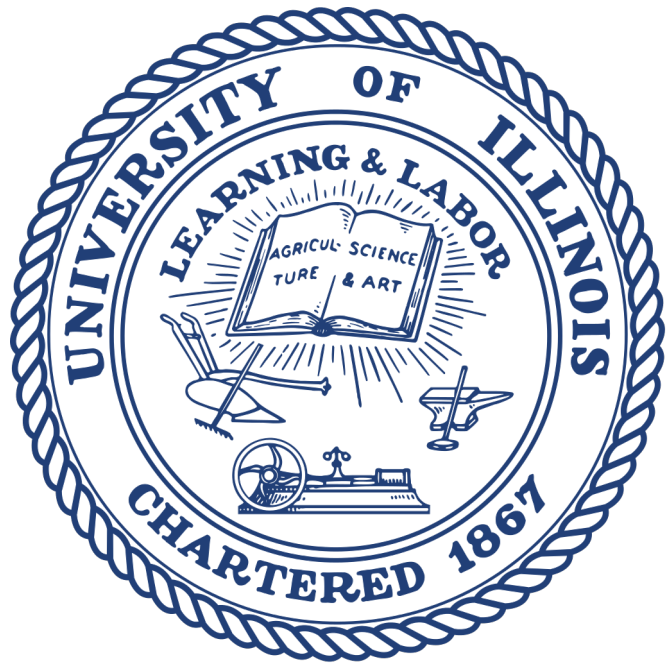

**Zixu Zhao**

Department of Electrical & Computer Engineering

University of Illinois at Urbana-Champaign

September, 2017

# Acknowledgements

The author of this report would like to express the deepest gratitude to Prof. Ahuja and his group for providing this fabulous research opportunity. This research is supervised by John M. Hart, Principal Research Engineer in Beckman Institute, and thanks are due to him for the thorough discussions on the technical problems and the outline of this research.

The author would like to acknowledge the great support and guidance received from Mr. Maziar Moaveni, who is a post-doctoral research associate in the Civil and Environmental engineering department of University of Illinois, for his kind guidance and sharp foresight of all the projects.

Finally, the author would like to recognize the time and efforts of all the research assistants spent providing ideas, data and other resources, a group that includes, but is not limited to, Benjamin L. Delay, Yifeng Chu and Haohang Huang.

The contents of this report only reflect the points of view of the author, who's also responsible for the facts and the accuracy of the data presented herein.

# Abstract

America has a massive railway system. As of 2006, U.S. freight railroads have 140,490 route-miles of standard gauge, but maintaining such a huge system and eliminating any dangers, like reduced track stability and poor drainage, caused by railway ballast degradation require huge amount of labor. The traditional way to quantify the degradation of ballast is to use an index called Fouling Index (FI) through ballast sampling and sieve analysis. However, determining the FI values in lab is very time-consuming and laborious, but with the help of recent development in the field of computer vision, a novel method for a potential machine-vison based ballast inspection system can be employed that can hopefully replace the traditional mechanical method. The new machine-vision approach analyses the images of the in-service ballasts, and then utilizes image segmentation algorithm to get ballast segments. By comparing the segment results and their corresponding FI values, this novel method produces a machine-vision-based index that has the best-fit relation with FI. The implementation details of how this algorithm works are discussed in this report.

Table of Contents

# 1. INTRODUCTION

## 1.1 BACKGROUND INFORMATION

Track ballast is packed under and between the ties and forms the trackbed upon which railroad ties are laid. It is used to bear the load from the railroad ties, to facilitate drainage of water, and also to keep down vegetation that might interfere with the track structure. It is observed in real life that ballast are subject to continuous abrasion and vibration as the trains pass through, and such vulnerability can cause severe degradation thus raising significant safety issues. The Fouling Index commonly used to quantify the degradation level is calculated by the summation of the percentage of the weight of ballast passing through a 4.75 mm sieve and a 0.075 mm sieve. [1][2]. To determine the percentage, this method involves sending highly experienced railway personnel to the field for visual inspection, selecting ballast samples and sieve testing in a laboratory. The accuracy depends on the experts' experience so it's affected by subjectivity. Also, this old sieving method suffers from low efficiency and locality, the new machine-vision based method captures the full-scale railway cross section and objectively calculates the degradation levels from these high-resolution images. In this way, the results have been observed to be easier and efficient. The new algorithm mainly includes watershed segmentation and image enhancements like gamma adjustment and image filtering, and other techniques like image thresholding, and their usefulness will be discussed in section two.

As the current research is underway and has made initial progress on developing an automated visual inspection system, such as mentioned in [2] and [3], this report aims to provide some implementation details on the currently successful Matlab version of this research using traditional computer vision methods, such as the reasons for choosing specific functions, a walkthrough of some important Matlab algorithms, as well as some tips and suggestions for modification of the algorithm. An automation plan and possible future work are also discussed at the end of this report. A new color coding technique is also used in this research to provide visual representation of the rock segments' sizes in order to make the fouling level more visually intuitive for the users.

# 2. SEGMENTATION ALGORITHM

## 2.1 ALGORITHM OVERVIEW

The core components of the proposed system as shown in the flow chart in Figure 1 includes three modules: a pre-processing module, a segmentation module and a post-processing module. In the pre-processing step, gamma and brightness adjustment is optional, which depends on the users' choices. Clear boundaries between the ballasts and a similar brightness level on the ballast surfaces have been observed to improve the segmentation results. The user can adjust the gamma and brightness level of ballast images in order to produce better particle segments. In our experiments, it's found that a gamma value of 1.45 and a brightness increase of 30% on an average darker image can produce better results. In Matlab a fast way to achieve this is by using the function "imhistmatch(input_img, sample_img)", in which the "sample_img" represents a sample ballast image that has an ideal level of gamma brightness values, and "input_img" represents the image whose brightness and gamma value are to be adjusted.

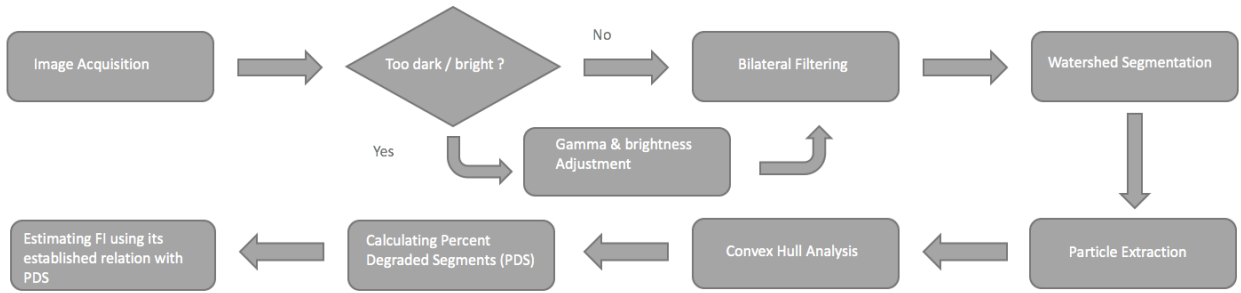

Fig.1. System Flow Chart

Unlike the typical smoothing filters like Gaussian Filter, which indiscriminately smoothes everything in an image, Bilateral Filter preserves the edges and reduces only the surface texture. Such a method is employed in our algorithm to remove the irregularities and edginess on ballast surface while preserving the edges between ballasts. Matlab doesn't offer a built-in Bilateral Filtering function, but a working code is not hard to find. In this work, we are citing the code from [4] based on the work [5] and [6].

After Bilateral Filtering, the images are ready to be processed at the segmentation stage. The assumption of watershed segmentation is that it assumes any greyscale image can be perceived as a topographic surface where the high intensity components represent the peaks and the low intensity components represent the valleys. The algorithm starts with filling these isolated valleys with "water" (color labels in the actual implementation), and as the water rises (color labels expand), color labels from different valleys eventually merge. The color labels become the outcome, each representing one segment.

In order to effectively avoid oversegmentation and remove the dark spots on the ballast surface, reconstruction-based opening and closing operations are used prior to the actual image segmentation. Reconstruction-based opening is erosion followed by morphological reconstruction, and reconstruction-based closing is dilation followed by morphological reconstruction [7]. This can be easily done in Matlab by calling "imerode" followed by "imreconstruct", and "imdilate" followed by "imreconstruct". After these operations, the regional maxima (the high intensity components representing the peaks) can be calculated using "imregionalmax" on the output image. Then a binary thresholded foreground can be used to calculate watershed ridge lines, that can be used as input for the Watershed Segmentation function.

The outputs of Watershed Segmentation are the extracted ballast particles segments, as shown in Figure 2. In the post-processing stage, each rock segment is generally expected to be convex, and a convex hull is constructed for each such segment (see Figure 3) to verify. The convexity test computes the ratio between the area of the segment and the area of the convex hull, and a segment should be labeled as non-ballast if the ratio falls below a certain threshold (we use 0.73 for such threshold but it's adjustable).

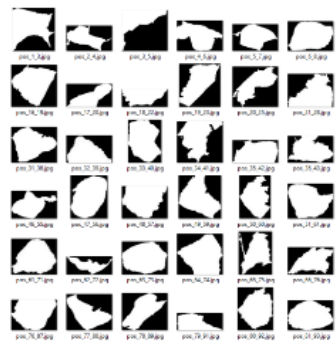

Fig.2. Ballast Segments after Segmentation

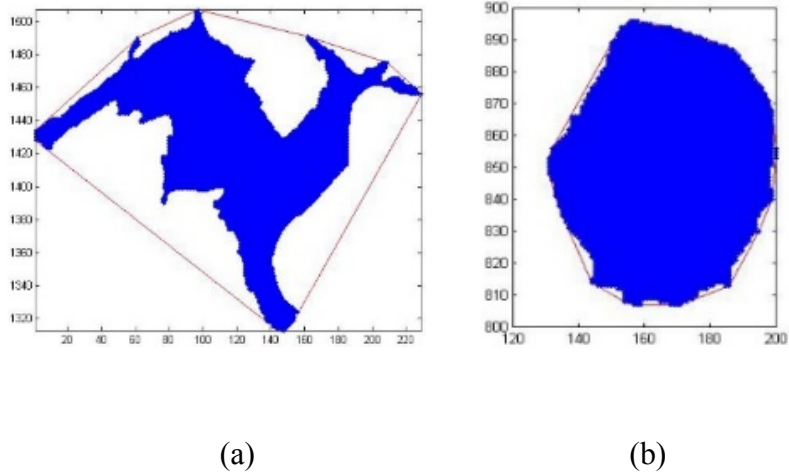

(a) (b)

Fig.3. (a) Convex hull (boundary) of a false-segment (non-ballast particle) and (b) Convex hull of a true-segment (ballast particle) – [Note: axes are pixel locations in the image]

The last step is degradation analysis on each rock segment. Firstly, we use an area-based approach to first identify the corresponding size categories for only the valid ballast particles that pass the Convex Hull Analysis. For this effort, the known area of a 1-inch (25.4 mm) diameter calibration ball, in terms of number of pixels, is used to estimate the area of each segment. The segments are then partitioned into three size categories including typical, small and large. The typical category represents average size ballast particles, the small category represents severely degraded particles, and the large represents oversized areas with particles too small to be identified individually, such as fine-grained soils. These categories are determined by normalizing the areas with respect to the area of the calibration ball and setting thresholds at less than 11% for small and more than 706.9% for large. The later categories are labelled as Degraded Segments.

A score for each image, i.e. Percentage of Degraded Segments (PDS), is defined as the percentage of the total area of large and small segments compared to the total area of the ballast image. Let $S_i$ ($i = 1,..., n$) be the areas of the segments $1,..., n$ in the image. Let $b$ be the area of the 1-in. (25.4 mm) diameter calibration ball and $A$ be the area of the image. Note that all areas are measured in terms of number of pixels. Then, the PDS value in terms of percentage can be computed using Equation 1 as follows:

Eq. 1:

$$Define: J = \{1 \leq i \leq n \mid threshold_s < \frac{s_i}{b} < threshold_l, threshold_s = 0.6, threshold_l = 3\}$$

$$Then, PDS(\%) = 100 \times (1 - \frac{\sum i \in J \, s_i}{A})$$

## 2.2   PARAMETER EXPLANATION

### 2.2.1   Spatial Gaussian Width and Range Gaussian Width

The bilateral filtering of an image at pixel position p relative to pixel positon q with intensity I is given in Equation 2 as follows:

$$\text{Eq.2.} \quad BF[I_p] = \frac{1}{w_p} \sum_{q \in s} G_{\sigma_s}(\|p-q\|) G_{\sigma_r}(|I_p - I_q|) I_q$$

where $1/w_p$ is the normalization factor, $G_{\sigma_s}$ and $G_{\sigma_r}$ are the Gaussian spatial kernel and the Gaussian range kernel. Equation 2 is a normalized weighted average where $G_{\sigma_s}$ is a spatial Gaussian that decreases the influence of distant pixels, $G_{\sigma_r}$ a range Gaussian that decreases the influence of pixels $q$ with an intensity value different from $I_p$. Consider the interested pixel at position $p$, the Spatial Gaussian width, represented by $\sigma_s$, determines the size (number of pixels) of the Gaussian window used to smooth images. In other words, Spatial Gaussian width is the size of the considered pixel neighborhood (thinner than $2\sigma_s$) used to calculate the average intensity of pixel at position $p$ [5]. Also, because bilateral filtering retains the boundaries of ballast particles, one needs the Range Gaussian width, represented by $\sigma_r$, to represent the minimum amplitude of an edge, so that the pixels in the considered neighborhood whose intensity is above or below this boundary are ignored in the smoothing process. In ballast images with 3158 × 4152 pixels, $\sigma_s$ and $\sigma_r$ could range from 3 to 200, but since some ballast boundaries before the pre-processing stage are already blurred, it is preferable to keep $\sigma_s$ below 20 and $\sigma_r$ below 10.

### 2.2.2   Strel-Size

The strel-size describes the radius (number of pixels) of a disk-shaped morphological structuring element used to clean up the ballast particles thus enhancing the segmentation results. It is found that in ballast images captured in the laboratory with 3158 × 4152 pixels, the strel-size needs to be 12 to 16 to segment 1-in. ballast particles, 16 to 20 for 2-in. particles, and 20 to 30 for 3-in. particles.

### 2.2.3   Number of Pixels for 1-in. (25.4 mm) Ball

The number of pixels for 1-in. (25.4 mm) calibration ball means how many pixels are needed to fill a 1- inch (25.4 mm) diameter of its area in the image. The user obtains this value at the pre-processing stage from the GUI.

### 2.2.4   Convex-hull Threshold

The convex hull of a set *X* of points in the Euclidean space is the smallest convex set that contains *X*. For instance, when *X* is a bounded subset of the plane, the convex hull may be visualized as the shape enclosed by a rubber band stretched around *X*. In the use of the algorithm, the convex-hull threshold falls between 0.6 and 0.75.

### 2.2.5   Graphical User Interface

Note that achieving desirable segmentation of each ballast image at a certain degradation level needs iterative interaction of user and the algorithm during the image processing task. This includes fine-tuning the segmentation parameters related to bilateral filtering, water segmentation as well as convex-hull threshold criteria in an effort to increase the accuracy of the segmentation. A trained user could do the processing in 2-3 minutes since the algorithm has only three sets of parameters. So, typically two or three iterations were required before achieving the desirable outcome. Generally, less than an hour is needed to train a user and it might take a couple of hours for the user to practice and get more efficient. Using the GUIDE tool in Matlab software, a Graphical User Interface (GUI) was developed to facilitate processing the captured images as well as providing a fast and efficient tool to monitor the effect of adjusting segmentation parameters on the processed images in each step. A screenshot of this GUI is shown in Figure 4.

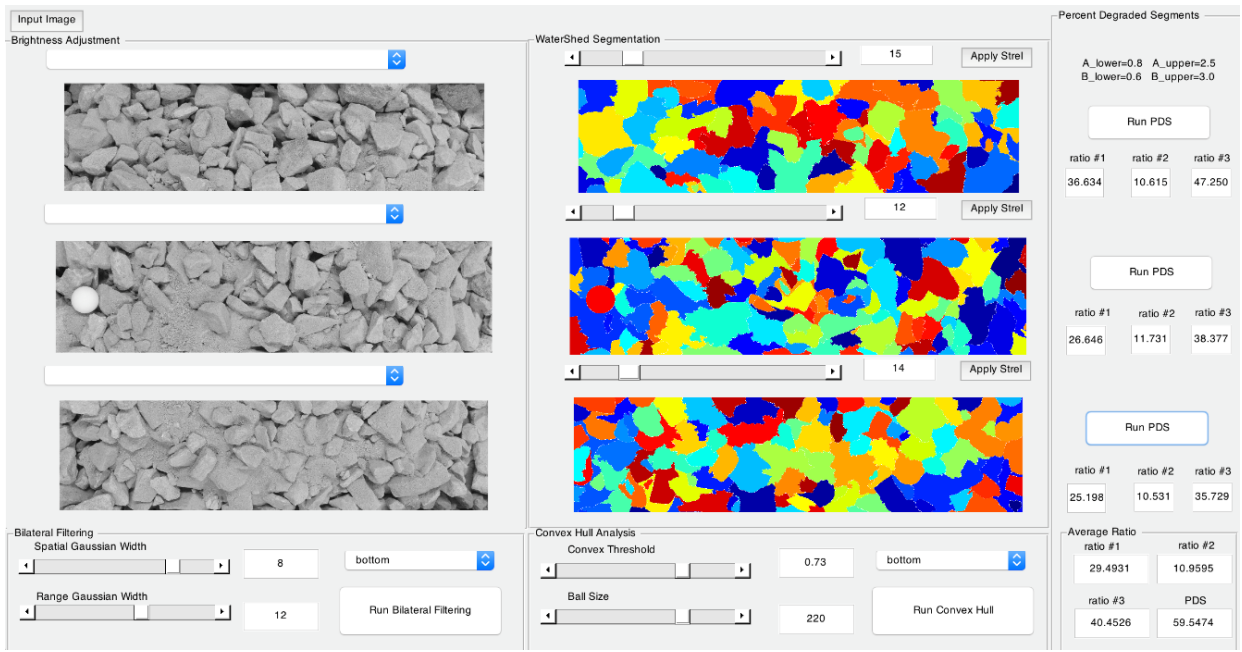

Fig.4. Graphical User Interface (GUI) for degradation evaluation of ballast images

## 2.3  PRODUCING CONSISTENT RESULTS

During the analyses of ballast images, it was observed that the performance of the segmentation algorithm in terms of precise detection of ballast particles and degradation zones was dependent on the level of particle size variability in each image. In other words, identifying one single set of segmentation parameters capable of detecting both small and large ballast particles for the entire image area was found to be very challenging. Consequently, it was decided that each ballast image would initially be cropped into three identical sub-images representing the upper, middle and lower portions of the captured ballast cross section. The reason for creating three sub-images can be explained by the expectation of increasing ballast degradation with increase in ballast layer depth, as observed previously for in-service track [8]. Therefore, the variability of particle sizes processed in each sub-image is expected to be less than the variability of particle sizes of the entire image. Through visual observation of the segmentation results, this approach proved to increase the performance of the algorithm. Thus, three sets of segmentation parameters and PDS values were recorded for each ballast image processed. The final PDS value for each image was reported as the average of the three PDS values associated with three sub-images.

Statistically speaking, the filtering and segmentation parameters for the bottom layer should be bigger than for the other layers, based on the observation that ballast size and coarseness in the bottom layer is obviously bigger than in the other two layers. However, in many cases the parameters in the bottom layer are at least as big as in the other two layers, or at most ten to twenty percent bigger than in the other layers. It is also preferred to keep the Bilateral Filtering parameters in all layers constant throughout the entire image dataset.

The success of Watershed Segmentation in the proposed system heavily depends on the strel-size and the evenness of brightness on ballast surfaces. The strel-size is usually set in a range of 10 to 20, depending on the general ballast size. The user can firstly try to set the strel-size at, for example, 14 on the first trial, and then tune the parameter higher or lower to try the second time. The user should keep tuning the parameters until an accurate segmentation result is generated. The relation of gamma and brightness to segmentation is also very important, but is tricky and requires strong intuition. Uneven brightness on the ballasts' surfaces can trick segmentation algorithm because it makes finding regional maxima harder (see section 2.1 for definition). Figure 5 offers an example of undersegmentaion (a) compared with proper segmentation (b), under the same parameters but with different brightness conditions in the original image.

A more graphical video tutorial of parameter tuning can be found at Youtube through this link [9].

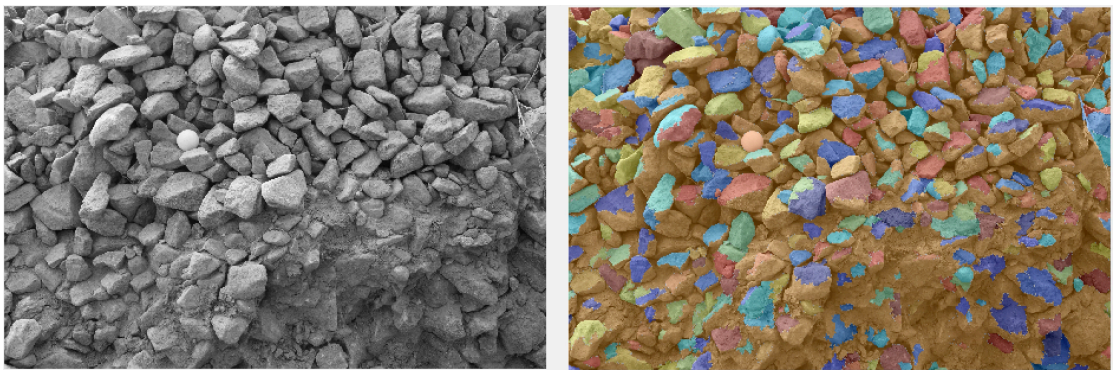

(a)
8

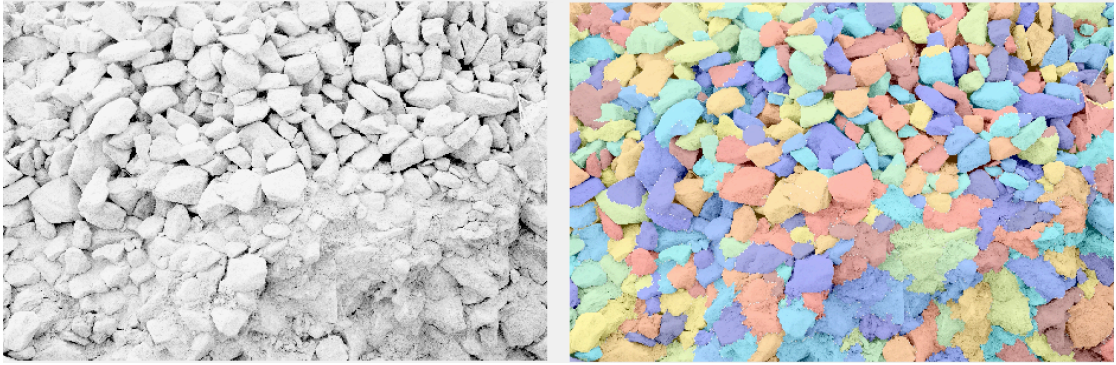

(b)

Fig.5. (a) unprocessed image with uneven brightness and its segmentation

(b) pre-processed image with even brightness and its segmentation

# 3. COLOR SIZE CODING KEY

## 3.1    COLORMAP DESIGN

For better visual evaluation of the segmentation results, a new color size coding key was added to the image segmentation module. As mentioned in section *2.1*, the segments are partitioned into three categories, typical, small and large, and the thresholds are set at 0.11 and 7.069 times the area of a 1-inch calibration ball. In Matlab implementation, a custom color coding key can be designed by calling "colormapeditor" function (see Figure 6).

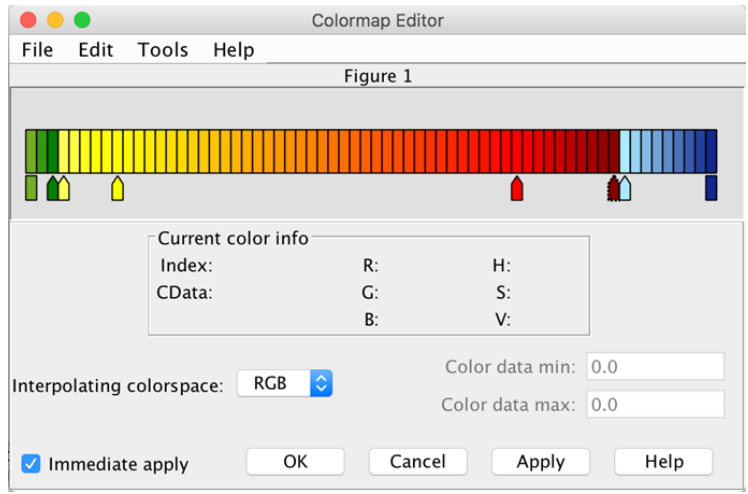

Fig.6. Matlab GUI for designing custom color coding key.

As indicated in Figure 7, detected small segments are labeled green color intensities while detected large segments are labeled blue color intensities. Note that these two, green and blue color categories, represent finer particles typically less than 3/4-in. and larger than 3-in. sizes of ballast particles, respectively. The segments that fall into the typical category are labeled using a color transition from yellow to red and continues to brown. Finally, the segments or degraded zones that did not pass the convex-hull test are labeled purple (indicated in Figure 6 as degraded zones). This new color size coding key is a desirable feature for assisting the users in understanding and fine-tuning the parameters.

This functionality has proved its huge usefulness in our recent experiments, which became less susceptible to user intuition and subjectivity since the color size coding key provided better visual results. We easily found problems in segmentation stage and degradation analysing stage. For example, the ballast partition threshold values mentioned above used to be 60% and 300%, which was responsible for many inaccurate experiment results. The causes of the inaccuracy remained undiscovered until the color size coding key was employed.

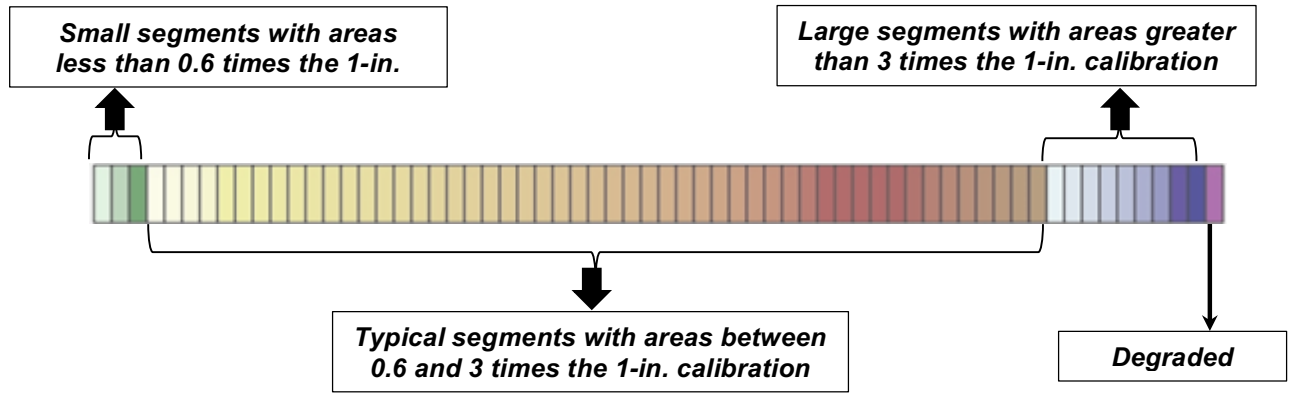

Fig.7. Color coded key to represent ballast image processing results considering sizes of segments

## 3.2 SOFTWARE MODIFICATION

### 3.2.1 Custom Color Coding in Place of Default Color Coding

To replace the default color coding key in the segmentation algorithm, a simple modification was employed. An algorithm with pseudo code is shown in Algorithm 1.

---

**Algorithm 1** Color Coding Key

---

**Input:** Input image for segmentation.

    Initialize Convex Threshold to 0.73.
    Label_matrix = WatershedSegmentaion(Input_img, strel_size)
    for each rock in the rock segments set:
        ratio = rock area / rock convex hull area
        if ratio < Convex Threshold:
            set segment pixels in Label_matrix to 0.
        end
        else:
            r = rock area / calibration ball area.
            if r < 0.11:
                set segment pixels in Label_matrix to ceil(r / 0.0367).
            end

```
            if 0.11 <= r < 7.069:
                set segment pixels in Label_matrix to ceil(r / 0.13645).
            end
            if 7.079 <= r < 11.569:
                set segment pixels in Label_matrix to ceil(r / 0.5).
            end
            if r >= 11.569:
                set segment pixels in Label_matrix to 64.
            end
        end

    end

    Convert label matrix to RGB image (using label2rgb function).

Output:  Color label matrix as an RGB image.
```

### 3.2.2 Stitch the Three Layers

As mentioned in *2.3*, all ballast images are divided into three sub-images for processing, then, quite often, ballast segments lying on the edge of the original image are also cut into two segments on separate sub-images, thus producing oversegmentation which inflates the fouling level. It is therefore a desirable functionality to stitch the three sub-images back again before the segments are extracted. The Algorithm 2 illustrates on reconstruction-based opening and closing operation, and Algorithm 3 modifies the Watershed Segmentation and Post-Segmentation, which eventually solves this issue.

**Algorithm 2** Modified Watershed Segmentation

**Input:** Input image after Bilateral Filtering.

1. I Perform image erosion followed by reconstruction.
2. Perform image dilation followed by reconstruction.
3. Find the regional maxima on each ballast and perform closing followed by an erosion.
4. Superimpose the regional maxima on the original image.
5. Calculate the watershed ridge lines.
6. Superimpose the ridge lines, regional maxima and the gradient magnitude on the original image.

**Output:** Image after reconstruction-based opening and closing operation, which is ready for watershed segmentation

| **Algorithm 3** Modified Watershed Segmentation and Post-Segmentation |
|---|
| **Input:** Perform the Algorithm 2 on the three layers respectively and get output images. |
| 1. Stitch the superimposed images from the three layers together.
2. Perform watershed on the stitched image and output a new label matrix.
3. Update the Label_matrix variable and perform Algorithm 1. |
| **Output:** Color coded segmented image |

Figure 8 shows a successful example of the segmentation results using the above three algorithms. The original image was first cut into three sub-images, and each layer went through Algorithm 1 and Algorithm 2. Then Algorithm 3 stitches the output images from the three layers together and uses Watershed Segmentation again to produce the image below.

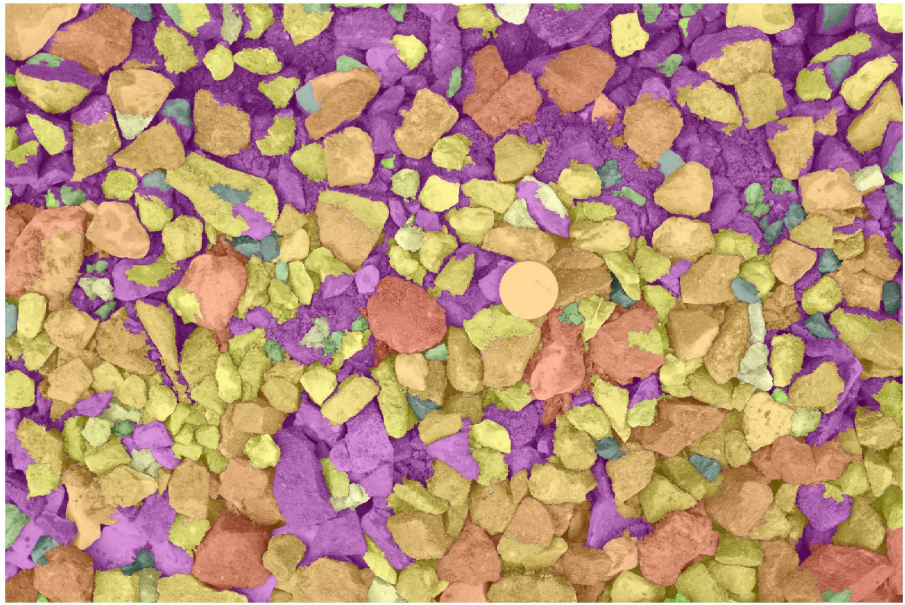

Fig.8. Color Coded Segmentation Results

# 4. AUTOMATION PLAN AND FUTURE WORK

## 4.1 PARAMETER AUTOMATION

To make the proposed system more automated, the five parameters, namely, the Range and Spatial Gaussian Width, strel-size, calibration ball size and Convex-hull Threshold should ideally not involve human decision and manual fine-tuning by the users.

The ball size and Convex-hull Threshold can be replaced by automation easily, because the need for ball size no longer exists if the camera is fixed at the same distance from the cross section, and the threshold has to remain the same for each image processed, and can't be changed through the entire project for consistency. The current threshold, 0.73, is a reasonable value and believed to produce accurate segments data. If more improvement is desired, a more ideal threshold value can be measured manually, or the entire Convex Hull Analysis process can be replaced by a simple machine learning algorithm trained on the data of segments' labels.

Spatial and Range Gaussian Width are not believed to have huge impacts on the segmentation results if they are just within a small range of variation. It's usually a common practice to keep the two parameters the same for the top and middle layers, and make the parameters for the bottom layer slightly higher. The preferred Spatial and Range Gaussian Width is 6 and 8 for the top and middle layers and 8 and 10 for the bottom layer. To make the proposed system automatically decide the parameters, it should base the parameter variation on the area of high and low frequency components of the image. It is believed that smooth surface has lower frequency variations and sharp edges, coarseness and fine particles have higher frequency variations. The system can increase the parameter if it detects larger high frequency area than normal and reduce the parameter if otherwise.

Tuning the strel-size requires very strong intuition and even a highly trained expert won't perform it well in an efficient manner. Intuition varies between different users, and can depend on many factors such as the ballast sizes, the overall fouling level, brightness, and so on. Figuring out the weights of these factors and how they relate to parameter variations can be quite challenging and susceptible to subjectivity. One of the possible ways to avoid such trouble and make this process more automated is the machine learning approach. A large dataset of the strel-sizes that are already determined by the users can serve as ground truth and their corresponding input images will train the system.

## 4.2 ALGORITHM IMPLEMENTATION

The proposed system is successfully implemented and finalized on Matlab, which offers a platform for testing and evaluation of the system. In the future, a possible implementation of this proposed visual inspection system on any machine can be done using C++ or Python. The image processing techniques like grayscale conversion, image cropping and resizing, brightness adjustment, bilateral filtering and Watershed Segmentation can be done using other packages like OpenCV.

# 5. CONCLUSION

As the previous studies have introduced a machine-vision based inspection system for ballast degradation levels in [2] and [3], this report offers more details on the implementation of the three main modules, pre-processing, segmentation, and post-processing, and the improvements made on the algorithms that had impacts on the final report [1]. Methods for operating the system and correlation of different sets of parameters are crucial for generating accurate and consistent lab data, so the insights and analysis in *2.3* helped a lot in generating data that was used in the final report [1]. The new color size coding key is another main topic discussed, and it helped to assist users in tuning the parameters and find problems in the system. Lastly, all the contents mentioned above help to define the imaging based index, Percent Degraded Segments (PDS), and establish significant correlation between Fouling Index (FI) and PDS values. With future implementation of the proposed system, it's promising that tedious and time-consuming ballast sampling and sieve analysis processes can be replaced. Accordingly, this innovative technology is envisioned for full development to provide automated means of track condition monitoring and installation on automated platforms and ballast shoulder cleaners to obtain degradation levels periodically along the entire path covered. Moreover, it can provide predictive ballast service life analyses and technology implementation in the management of ballast maintenance.